\documentclass[review]{elsarticle}

\usepackage{lineno,hyperref}
\usepackage{booktabs}
\usepackage{url}

\journal{Journal of \LaTeX\ Templates}

\begin{document}

\begin{frontmatter}

\title{Video-based Smoky Vehicle Detection with A Coarse-to-Fine Framework}

\author{Xiaojiang Peng$^{a}$, Xiaomao Fan$^a$, Qingyang Wu$^a$, Jieyan Zhao$^b$, Pan Gao$^c$}
\address{$^a$Shenzhen Technology University, Shenzhen, P.R. China, 518118}
\address{$^b$Central South University, Changsha, P.R. China, 410083}
\address{$^c$Nanjing University of Aeronautics and Astronautics, Nanjing, P.R. China, 211106}




\begin{abstract}
Automatic smoky vehicle detection in videos is a superior solution to the traditional expensive remote sensing one with ultraviolet-infrared light devices for environmental protection agencies. However, it is challenging to distinguish vehicle smoke from shadow and wet regions coming from rear vehicle or clutter roads, and could be worse due to limited annotated data.
In this paper, we first introduce a real-world large-scale smoky vehicle dataset with 75,000 annotated smoky vehicle images, facilitating the effective training of advanced deep learning models. To enable fair algorithm comparison, we also build a smoky vehicle video dataset including 163 long videos with segment-level annotations. Moreover, we present a new Coarse-to-fine Deep Smoky vehicle detection (CoDeS) framework for efficient smoky vehicle detection. The CoDeS first leverages a light-weight YOLO detector for fast smoke detection with high recall rate, and then applies a smoke-vehicle matching strategy to eliminate non-vehicle smoke, and finally uses a elaborately-designed 3D model to further refine the results in spatial temporal space. Extensive experiments in four metrics demonstrate that our framework is significantly superior to those hand-crafted feature based methods and recent advanced methods. The code and dataset will be released at https://github.com/pengxj/smokyvehicle.


\end{abstract}


\end{frontmatter}


\section{Introduction}
Motor vehicles are the main source of urban carbon monoxide, hydrocarbons, and oxides of nitrogen, that are responsible for the formation of photochemical smog~\cite{science1995}. In the world, especially in developing countries, fuel vehicles produce a large number of emissions which affect our health~\cite{2009Real}. Smoky vehicles refer to those vehicles with visible black smoke exhausted from vehicle exhaust hole. Automatic smoky vehicle detection is very useful for controlling emission, endowing environmental protection agencies with smart surveillance.

Generally, there are several techniques for smoky vehicle detection in industry. An old type of smoky vehicle detection method works manually which includes public reporting, regular road inspection and night
inspection by the law enforcement workers, and manual video monitoring.
Recent advanced methods including remote sensing vehicle exhaust monitoring with ultraviolet-infrared light, and automatic detection in surveillance videos~\cite{TAO202017}. Due to the cost of remote sensing with ultraviolet-infrared light is extremely expensive, detecting smoky vehicles in surveillance videos has attracted increasing attention recently~\cite{Tao2018,Cao2018,Tao2019b,Tao2019a,Cao2019,Tao2019,Tao2020,Wang2021} which essentially is to detect the visible black smoke and vehicles.

\begin{figure}[t]
    \centering
    \includegraphics[width=0.3\textwidth,height=0.3\textwidth]{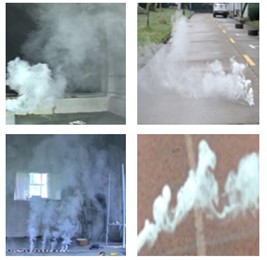}~\includegraphics[width=0.3\textwidth,height=0.3\textwidth]{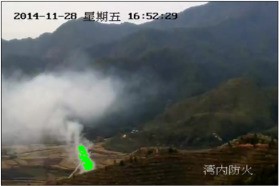}~\includegraphics[width=0.4\textwidth,height=0.3\textwidth]{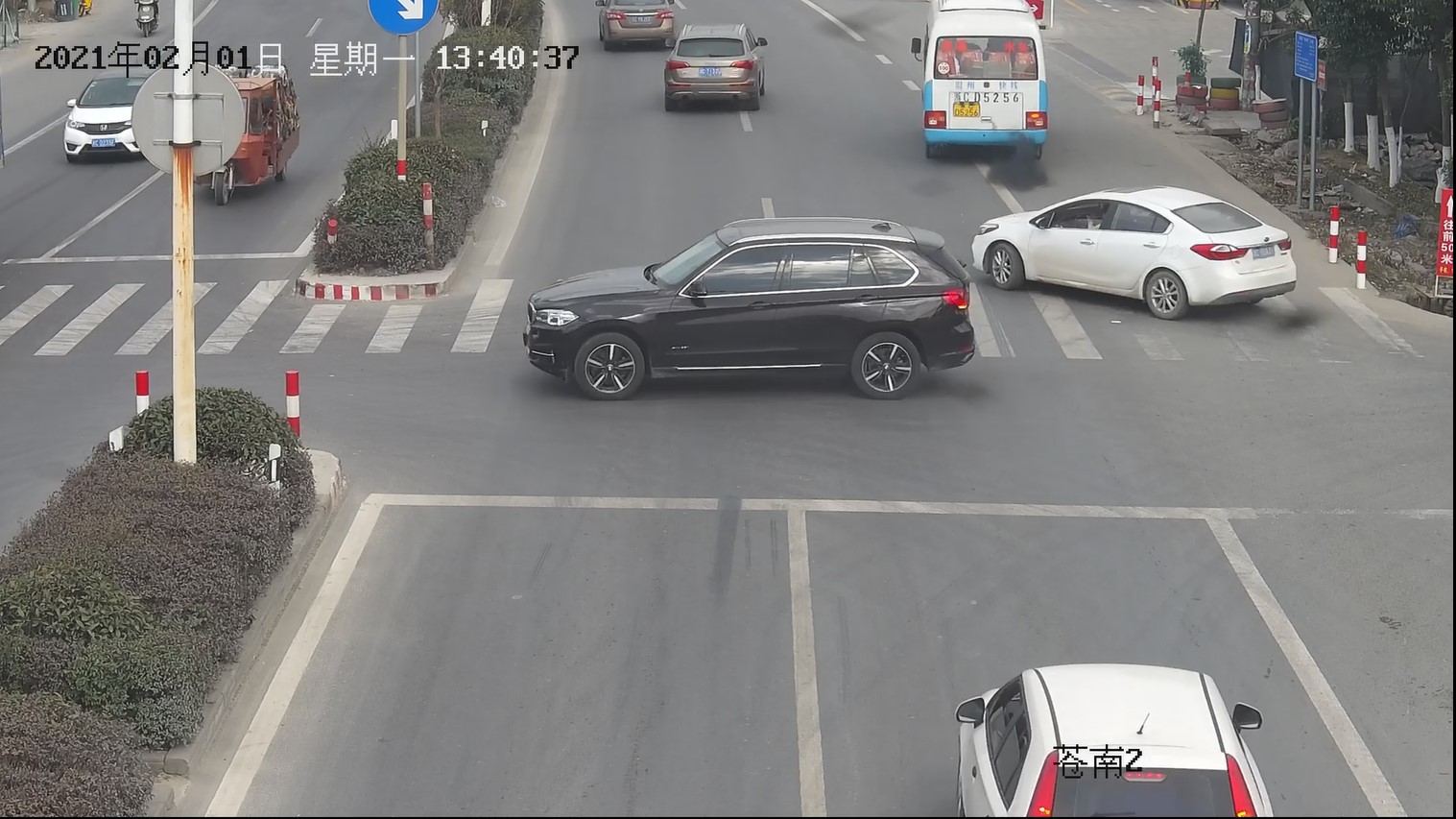}
    \caption{Smoke detection in different environments. Left: general wildfire smoke~\cite{2017Smoke}. Middle: forest wildfire smoke~\cite{ZHOU201650}. Right: vehicle smoke on road (ours).}
    \label{fig:smokecompare}
\end{figure}

In the past two decades, smoke detection and recognition in videos has been studied extensively~\cite{toreyin2005wavelet,xiong2007video,yuan2008fast,ZHOU201650,2017Smoke,yuan2018learning,yuan2019encoding,2019Convolutional,yuan2019fusing,2020Encoding}, and most of works mainly focus on wildfire scenario while few attentions are paid on the road vehicle scenario. For a long time texture features was dominated in smoke recognition. For example, wavelet features are used in \cite{toreyin2005wavelet} and local binary patterns are used in \cite{YUAN20124326,ZHOU201650,2017Smoke,yuan2019encoding,yuan2019fusing}. Considering the motion information of smoke, motion orientation features~\cite{yuan2008fast}, dynamic texture features~\cite{2017Smoke}, and 3D local difference features~\cite{yuan2018learning} are also widely used. After feature extraction for regions, a classifier like Support Vector Machine (SVM)~\cite{1998Statistical}, is used to categorize each region as smoke and non-smoke region. Recently, with the development of deep learning~\cite{2012ImageNet}, convolutional neural networks have been extensively used for smoke recognition~\cite{2017Comparative,20183D,2019Smoke,2019Deep,2019Convolutional,2019Recurrent,Yaocong2018Real,2018Fire}. \cite{2017Comparative} have evaluated different convolutional neural networks for image-based smoke classification including AlexNet~\cite{2012ImageNet}, ResNet~\cite{2016Deep}, VGGNet~\cite{2014Very}, etc. 3D convolutional neural networks are applied for smoke sequence recognition in \cite{20183D,Yaocong2018Real,2019Smoke}. Recurrent networks are also explored for sequence-based smoke recognition in \cite{2019Recurrent}. Almost all the mentioned methods focus on general wildfire smoke or forest wildfire smoke as shown in the left and middle of Figure \ref{fig:smokecompare}.

Intuitively, the above smoke recognition methods can be used for smoke vehicle detection without considering vehicles. Along this direction, Tao \textit{et al.}~\cite{Tao2019a} leveraged a codebook model for background modeling and applied volume local binary count patterns, which achieved impressive smoke detection results in strong smoky traffic videos. However, there exists essential differences between wildfire smoke and vehicle smoke as shown in Figure \ref{fig:smokecompare}. First, the smoke emitted from vehicles is mostly slighter than wildfire smoke, and occupies only a small region in video frames. Second, due to smokes are emitted from the bottom of vehicles, algorithms are apt to confuse smokes, shadows, raining road surfaces. Though large progresses have been made in recent years\cite{Tao2018,Cao2018,Tao2019b,Tao2019a,Cao2019,Tao2019,Tao2020,Wang2021}, the false alarm rate over frames is still high (e.g., around 13\% in \cite{TAO202017}) which is hard to meet practical applications. Worse still, unlike the wildfire smoke recognition, there is no public available vehicle smoke data in the community. The above issues make smoky vehicle detection more challenging and is still an open problem.

In this paper, to address these challenges, we first introduce a real-world Large-Scale Smoky Vehicle image dataset, termed as LaSSoV, which is the largest vehicle smoke dataset up to date. To build the LaSSoV, we hire trained staffs to carefully check a mount of surveillance videos and annotate smoke regions every 10 frames if existing. Totally, the LaSSoV contains 75,000 smoky vehicle images with bounding box annotations, which is expected to facilitate the effective training of deep learning models in smoky vehicle detection. To simulate practical scenarios, we also provide a smoky vehicle video dataset, termed as LaSSoV-video, which includes 163 segment-annotated long videos in diverse environments.

Second, we propose a Coarse-to-fine Deep Smoky vehicle detection (CoDeS) framework which is composed of a light-weight YOLO~\cite{2016You} based smoke detector, a smoke-vehicle matching module, and a light-weight 3D convolutional network (3DCNN). The light-weight smoke detector, called YOLOv5tiny, is improved from YOLOv5n\footnote{https://github.com/ultralytics/yolov5} by replacing most of the convolutional operations with depth-wise and pointwise convolutional operations inspired by MobileNet~\cite{2018MobileNetV2} and GhostNet~\cite{2020GhostNet}. It is more efficient than the official released YOLOv5n model. The YOLOv5tiny can quickly detects smoke regions in each video frame yet may be confused by shadow and non-road regions. The smoke-vehicle matching module is activated once a smoke region is detected in a frame. This module first runs a vehicle detector to get vehicle regions, 
and then computes the Intersection-over-Union (IoU) ratios or pair-wise distances between smoke regions and vehicle regions. If a certain vehicle owns the minimum distance which is smaller than a given threshold or owns the largest IoU, we make the smoke region and the certain vehicle as a smoke-vehicle pair. The smoke-vehicle matching module can remove these false positives easily like non-road and other non-vehicle regions. 

As found in practice, the detected regions of static images from smoke detector are difficult for human beings to further check if they are cropped, while we can easily distinguish them by watching the spatial-temporal extent of each region. Inspired by this fact, to further remove hard false positives, we consider the spatial-temporal extent of a smoke region, and develop a light-weight 3DCNN to classify the spatial-temporal extent for a refinement. Specifically, we explore two kinds of short-term 3DCNN based on ResNet18: one is called 3DCNN-prefix which replaces the early 2D convolutional layers of ResNet18 with 3D ones, and the other is called 3DCNN-suffix which performs 3D convolution in the last feature maps of ResNet18 over several frames.

Finally, we extensively evaluate our method and other recent advanced ones in four frame-wise metrics including detection rate, false alarm rate, precision, and F1 score. Almost all previous works only provide one or two of them for performance evaluation, which are not fair for comprehensive comparison.
We conduct experiments on our LaSSoV and LaSSoV-video since there is no public smoky vehicle data, and demonstrate that our framework is significantly superior to those hand-crafted feature based methods and recent advanced methods. 

In the remaining of this paper, we briefly review the related works on smoke detection and smokey vehicle detection in Section \ref{relate}. We then introduce our datasets and compare them to existing datasets in Section \ref{dataset}. We further describe our methods in Section \ref{approach} and present experimental results in Section \ref{experiment}. Finally, we give a conclusion and discuss the future work in Section \ref{conclusion}.

\section{Related work}
\label{relate}
Smoky vehicle detection is a challenging scientific and engineering problem. We briefly review the related works on wildfire visual smoke detection and recent smoky vehicle detection in this section.

\subsection{Visual smoke detection}
Visual smoke detection aims to detect fire by recognizing early smoke or fire smoke anywhere within the field of the camera at a distance. In the past two decades, video based wildfire smoke detection has been extensively studied. Here, we roughly category the video-based smoke detection methods into shallow feature based methods and deep learning based methods.

\textit{Shallow feature based methods}. 
These methods mainly extract structural and statistical features from visual signatures such as motion, colour, edge, obscurity, geometry, texture and energy of smoke regions. 
Guillemant and Philippe~\cite{Guillemant2001Real} implemented a real-time automatic smoke detection system for forest surveillance stations based on  the assumption that the energy of the velocity distribution of smoke plume is higher than other natural occurrences.
Gomez-Rodriguez \textit{et al.}~\cite{2002Smoke} used optical 
flow and wavelet decomposition algorithm for wildfire smoke detection and monitoring. T{\"o}reyin \textit{et al.}~\cite{toreyin2005wavelet} presented a  close range (in 100 meters) smoke detection system which is mainly comprised of moving region detection and spatial-temporal wavelet transform. Yuan~\cite{yuan2008fast} proposed an accumulative motion orientation feature based method for video smoke detection. Lin \textit{et al.}~\cite{2017Smoke} applied volume local binary patterns for dynamic texture modelling and smoke detection. Yuan \textit{et al.}~\cite{yuan2019encoding} considered the scale invariance of local binary patterns and utilized hamming distances for local binary patterns to address image-based smoke recognition. Yuan \textit{et al.}~\cite{2020Encoding} leveraged multi-layer Gabor filters for image-based smoke recognition. Overall, all these shallow feature based methods focused on improving the robustness of hand-crafted texture or motion features which may be limited in their representation abilities. 

\textit{Deep learning based methods}.
With the development of deep learning~\cite{2012ImageNet}, convolutional neural networks (CNNs) have been extensively used for visual image classification and object detection recently~\cite{2016Deep,2016You}.
Visual smoke detection can be viewed as an object classification or detection task. From the characteristic of deep networks, we can roughly divide these deep networks into 2D appearance networks and spatial-temporal networks.

For 2D networks, Frizzi \textit{et al.}~\cite{2016Convolutional} built their  CNNs very similar to well-known LeNet-5~\cite{1998Gradient} with an increased number of feature maps in convolution layers, and used real smoke scenarios and achieved 97.9\% accuracy which is higher than
the performance of traditional machine learning approaches.
Tao \textit{et al.}~\cite{2016Smoke} applied the well-known AlexNet~\cite{2012ImageNet} for visual smoke detection and achieved high performance on open their smoke image dataset.
Filonenko \textit{et al.} provided a comparative study of modern CNN models.
To address image-based smoke detection, Gu \textit{et al.}~\cite{2019Deep} designed a new deep dual-channel neural network for high-level texture and contour information extraction.

For spatial-temporal networks, Lin \textit{et al.}~\cite{2019Smoke} designed a 3D CNN to extract the spatial-temporal features of smoke proposals. Tao \textit{et al.}~\cite{Tao2019} built a professional spatial-temporal model which is comprised of three CNNs on different orthogonal planes. Li \textit{et al.}~\cite{20183D} proposed a 3D parallel fully convolutional networks for real-time video based smoke detection. In the other direction, recurrent neural networks are also applied for video-based smoke detection \cite{2019Recurrent}. 

Overall, almost all the outlined methods focus on general wildfire smoke recognition or detection.

\subsection{Smoky vehicle detection}
Smoky vehicle detection includes smoke detection and vehicle assignment, aiming to capture those vehicles that are emitting visible smoke. The obvious difference between general wildfire smoke and vehicle smoke is that the backgrounds of vehicle smoke are varied roads which contain vehicle shadows, dirty regions, etc.

One direction of smoky vehicle detection only considers to detect vehicle smoke regions as accurate as possible, which can be resorted to general smoke detection methods. Tao \textit{et al.}~\cite{Tao2019a} leveraged a codebook model for background modeling and applied volume local binary count patterns, which achieved impressive smoke detection results in strong smoky traffic videos. Tao \textit{et al.}~\cite{Tao2018} also extracted the multi-scale block tamura features for smoke proposals and classification. To avoid reshaping smoke regions and keep the shape of smoke plume, Cao \textit{et al.} developed a spatial pyramid pooling convolutional neural network (SPPCNN) for smoky vehicle detection. 

The other direction of smoky vehicle detection jointly considers to detect smoke regions and vehicles. Tao \textit{et al.}~\cite{Tao2019b} used hand-crafted features to jointly detect rear vehicle and smoke.
Cao and Lu~\cite{Cao2019} first leveraged background subtraction to detect motion rear vehicles, then applied Inception-V3~\cite{szegedy2016rethinking} and a temporal Multi-Layer Perception (MLP) for smoke classification. \textit{In practice, a background modelling method like Gaussian Mixed Model (GMM) can be more expensive than a optimized YOLOv5 in computational cost}. Wang \textit{et al.}~\cite{Wang2021} resorted to an improved YOLOv5 model for fast smoky vehicle detection which actually only detects smoke regions.
Different from all the above methods, our coarse-to-fine smoky vehicle detection framework first leverages a optimized light-weight YOLO model to capture smoke region proposals, then utilizes a smoke-vehicle matching module to remove a certain of false positives, and finally uses a light-weight 3DCNN classification model to further refine.

\section{Smoky vehicle datasets}
\label{dataset}
In the computer vision community, there exist several widely-used wildfire smoke datasets which have promoted the progress of fire and smoke recognition, e.g., KMU Fire-Smoke~\cite{Ko2012Wildfire}, Mivia Fire-Smoke~\cite{csvt2015_fire}, and VSD~\cite{YUAN2016225}. However, up to date, there is no publicly-available smoky vehicle datasets. In deep learning era, a good qualitative dataset is crucial for algorithm performance. To this end, we collect a large-scale smoky vehicle image dataset for model training and another video dataset for algorithm comparison.

\begin{figure}[t]
    \centering
    \includegraphics[width=\textwidth]{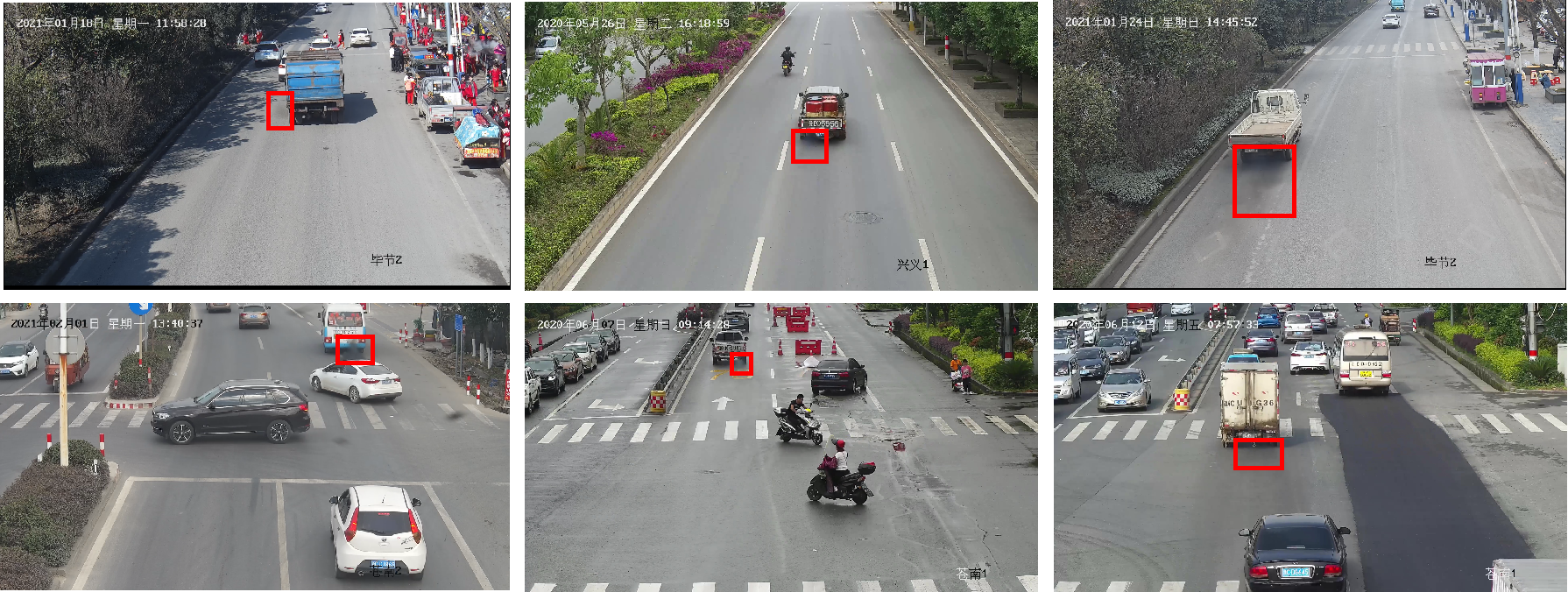}
    \caption{Some samples in our LaSSoV dataset. Note that our dataset includes light smoke frames, strong smoke frames, as well as crossroad smoke frames.}
    \label{fig:lassovsamples}
\end{figure}

\subsection{Dataset collection and annotation}
To manually collect a large-scale smoky vehicle image dataset from scratch is pretty hard since there may only exists few smoky frames in one day's surveillance footage. To this end, we make a cold boot process for data collection. First, we rent several smoky vehicles and drive them in the view of different surveillance cameras. Then, we capture all the smoky frames and use the LabelMe toolbox\footnote{https://github.com/CSAILVision/LabelMeAnnotationTool} to annotate smoke bounding boxes. Specifically, we mainly get 1,000 smoke frames at this stage. After that, with the annotated smoke images, we manage to train a rough YOLOv5s smoke detector. Finally, we deploy this detector in several surveillance cameras and gather these frames with smoke detection threshold 0.05. 

With the cold boot process, we manage to collect thousands of samples every day. To mine real smoky vehicle frames, we hire several people to carefully check whether the collected frame has smoke region or not. Specially, to make the verification easy, the screen is adapted to resolution 1920$\times$1080 since all images are 1920$\times$1080. If a frame contains smoke, the people is asked to annotated the smoke bounding box using LabelMe. Within several months, we have annotated 75,000 smoke frames totally. We call this LArge-Scale SmOky Vehicle image dataset as LaSSoV. The LaSSoV includes light smoke frames, strong smoke frames, as well as crossroad smoke frames. Figure \ref{fig:lassovsamples} illustrates several samples in the LaSSoV.


\textit{LaSSoV-video.} Since temporal information is important for smoke recognition and existing private smoky vehicle datasets are comprised of videos, we further collect long-term smoke videos for temporal model exploration and performance comparison. We call the data as LaSSoV-video. In total, we collect 163 one-minute videos with vehicle smoke, and use PotPlayer\footnote{http://potplayer.daum.net/} to annotate the start smoke frame and the end smoke frame. 
Specifically, since we may use videos for temporal model training, we randomly split LaSSoV-video into two sub sets, i.e., a training set which includes 103 videos and a testing set which contains 60 videos.

\begin{table}[t]
\caption{The comparison between existing datasets and ours.}
\begin{tabular}{@{}llllll@{}}
\toprule
Database & Type & Total samples & Annotation  & Available \\ \midrule
   KMU Fire-Smoke~\cite{Ko2012Wildfire}      &  video    &   10     &    category     &     public       \\
   Mivia Fire-Smoke~\cite{csvt2015_fire}     &   video   &     149   &   category         &        public    \\
   SmokeVideos in \cite{2017Smoke}      &   video   &     25   &     block labels         &   public   \\
VSD(Set4)~\cite{YUAN2016225}     &  patches  &  10,617  &   category   &  public         \\
\midrule
SEU-SmokeVeh\cite{Tao2019a}     &  videos  &  102  &   temporal segments   &  private         \\
 \textbf{LaSSoV}(Ours)     &   image   &    75,000         &      bounding boxes        &  public        \\
 \textbf{LaSSoV-video}(Ours)     &   video   &    163       &      temporal segments        &  public        \\\bottomrule
\end{tabular}
\label{tab:datacomparison}
\end{table}

\subsection{Dataset comparison}
\begin{figure}[t]
    \centering
    \includegraphics[width=\textwidth]{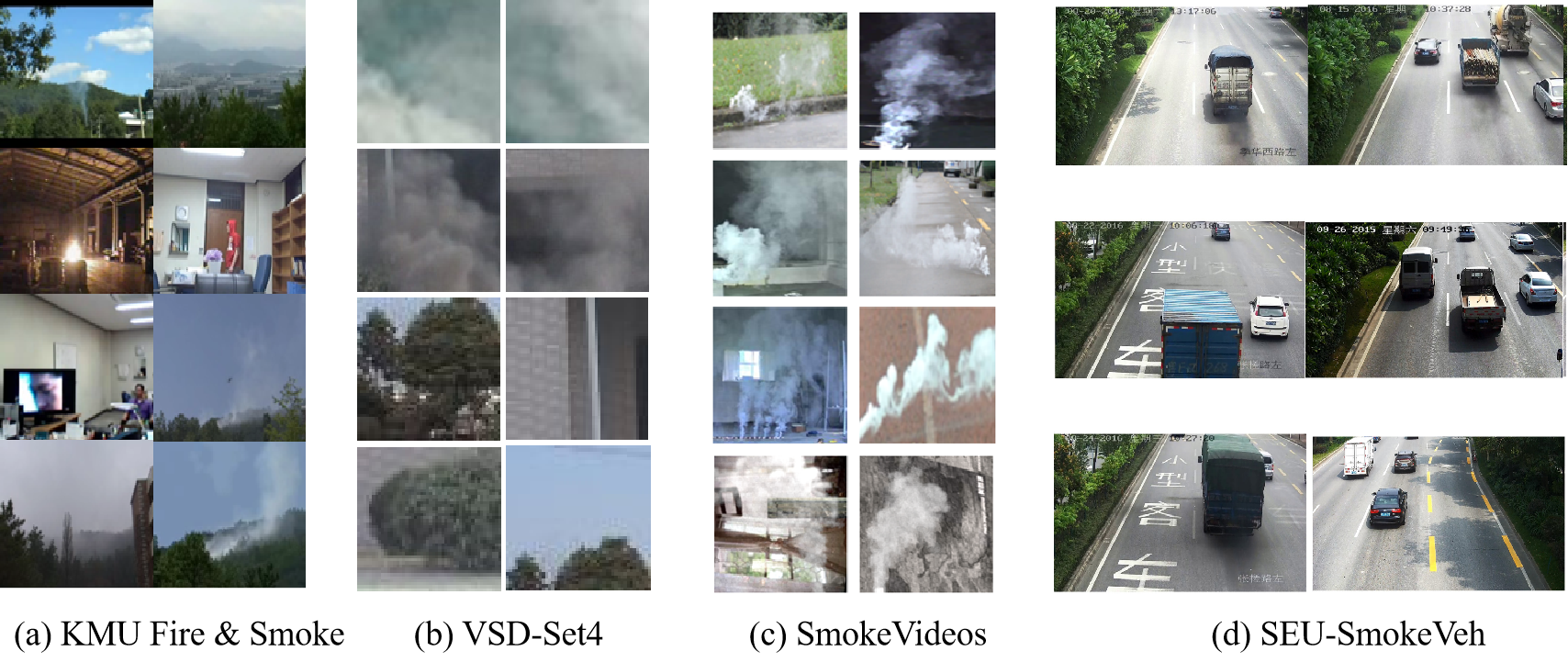}
    \caption{Some examples of existing datasets.}
    \label{fig:smokedatasets}
\end{figure}

We compare our datasets to several existing datasets in Table \ref{tab:datacomparison}. The public KMU Fire-Smoke dataset~\cite{Ko2012Wildfire} is collected by the KMU CVPR Lab which includes indoor and outdoor flame, indoor and outdoor smoke, wildfire smoke, and smoke or flame-like moving objects. Figure \ref{fig:smokedatasets}(a) shows some frames in the KMU Fire-Smoke dataset. The Mivia Fire-Smoke dataset~\cite{csvt2015_fire} contains 149 long-term smoke videos which are mainly captured in wildfire environment with the occurrence of fog, cloud, etc. The smoke video dataset in \cite{2017Smoke} (see Figure  \ref{fig:smokedatasets}(c)) is comprised of 25 short distance smoke videos. Its block annotations can be used for local feature based algorithms. The VSD(Set4)~\cite{YUAN2016225} is built from similar videos in \cite{2017Smoke} while all the smoke patches and random non-smoke patches are extracted for smoke recognition, see Figure \ref{fig:smokedatasets}(b). Though these datasets have promoted the progress of smoke recognition and detection, they are quite different from on-road smoke emitted from vehicles.

Perhaps the most similar dataset with ours is the private SEU-SmokeVeh\cite{Tao2019a} dataset, which contains 98 short video and 4 long videos. Each short video has one smoke
vehicle, and each long video has many non-smoke vehicle and
1--3 smoke vehicles. The videos totally have 5937 smoke frames
and 151,613 non-smoke frames. All frames in SEU-SmokeVeh are downsampled to
the size of 864 × 480 pixels. For our LaSSoV-video, it includes 163 high-resolution videos with 12,287 smoky vehicle frames and more than 100,000 non-smoke frames. Moreover, the LaSSoV-video covers 4 scenes with different weather. In addition, since in deep learning era, training data is the devil of good performance, we also provide LaSSoV which is the largest image smoky vehicle datasets so far. It is expected to promote the progress of smoky vehicle detection.

\section{Approach}
\label{approach}

In this section, we first provide an overview of our Coarse-to-fine Deep Smoky vehicle detection  (CoDeS) framework, and then present the details of its modules, i.e. the light-weight detector, the smoke-vehicle matching module, and the light-weight 3DCNN model.

\begin{figure}[t]
    \centering
    \includegraphics[width=0.9\textwidth]{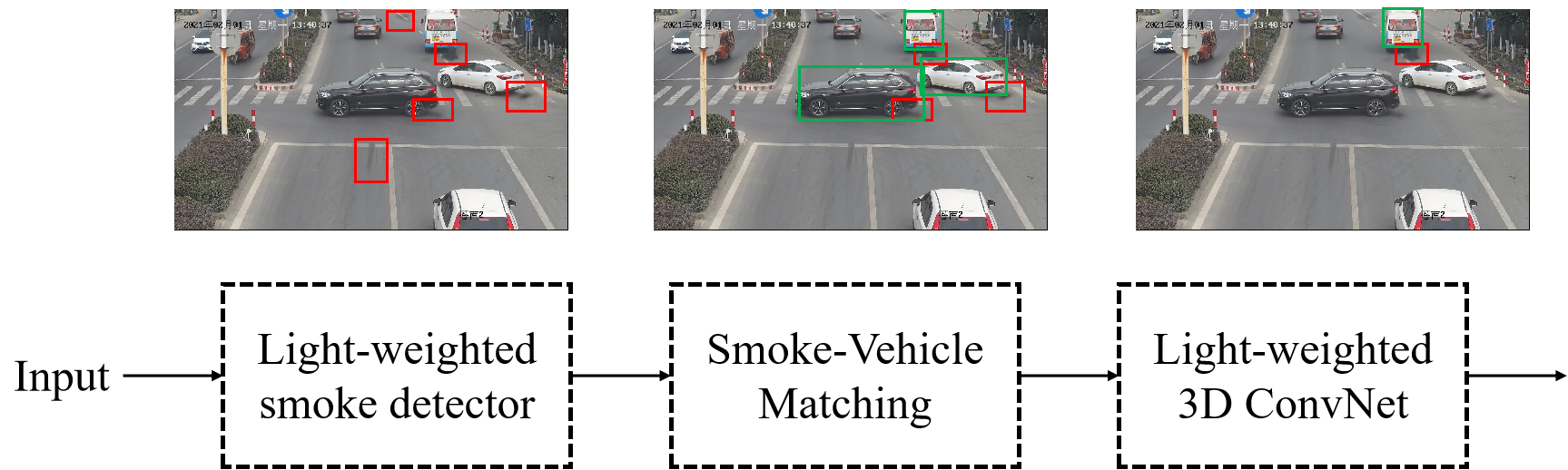}
    \caption{The pipeline of our coarse-to-fine deep smoky vehicle detection  (CoDeS). The output of each module is illustrated in the top images.}
    \label{fig:framework}
\end{figure}
\subsection{Overview of our framework}
Due to the lack of smoky vehicle data, there are not any \textit{pure} deep learning algorithms for smoky vehicle detection. Instead, background modelling and proposal detection are normally used in existing methods. Here we propose a Coarse-to-fine Deep Smoky vehicle detection  (CoDeS) algorithm without traditional pre-processing steps, which is illustrated in Figure \ref{fig:framework}. 

The framework of our CoDeS is comprised of three modules, namely a light-weight smoke detector, a smoke-vehicle matching module, and a light-weight 3D convolutional network. 
The light-weight smoke detector is actually an improved YOLOv5n detector, which largely uses depth-wise and point-wise convolutional operations to replace traditional convolutional operations. 
The light-weight smoke detector detects smoke regions in each video frame and may be confused by shadow and non-road regions since their appearances are very similar with vehicle smoke, see Figure \ref{fig:framework}. The smoke-vehicle matching module is activated once a smoke region is detected in a frame. This module first runs a vehicle detector to get vehicle regions. To this end, we roughly use the off-the-shelf (on COCO dataset) YOLOv5n which can detect three vehicle categories, \textit{i.e.} car, bus, and truck. With the assumption that each smoke region corresponds to a vehicle, this module then computes the Intersection-over-Union (IoU) ratios and pair-wise distances between smoke regions and vehicle regions. If the minimum distance is smaller than a given threshold or the IoU is larger than a threshold, we make the given smoke region and the certain vehicle as a smoke-vehicle pair. The smoke-vehicle matching module can remove these false positive regions easily like non-road and other non-vehicle regions, see Figure \ref{fig:framework}. To further remove hard false positives, another light-weight 3DCNN model is applied on the spatial-temporal extent of the remaining regions. Overall, the CoDeS is accurate and efficient which largely leverages deep learning technology.

\begin{figure}[t]
    \centering
    \includegraphics[width=\textwidth]{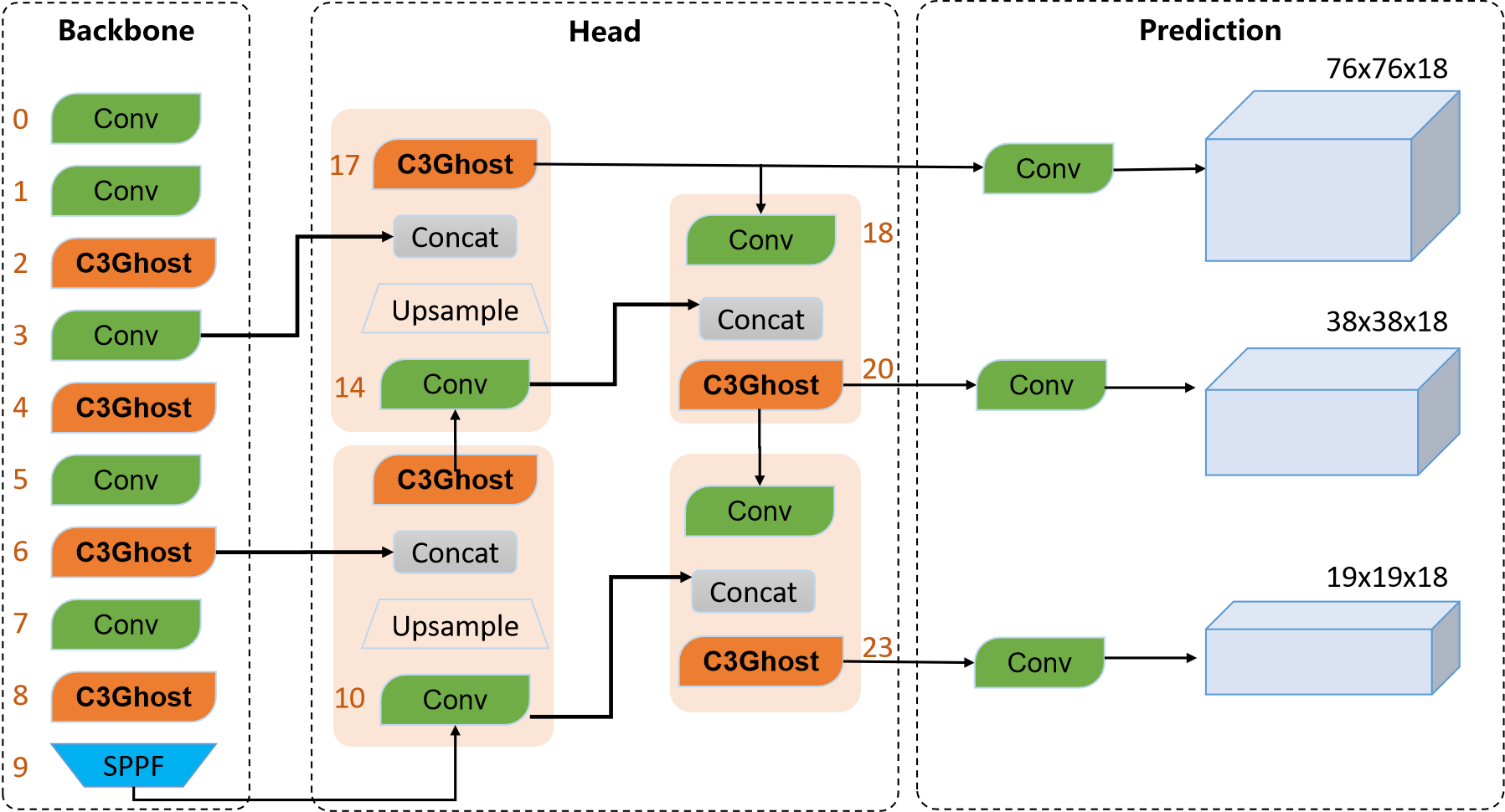}
    \caption{Our light-weight YOLOv5 vehicle smoke detector, termed as YOLOv5tiny. It is modified from the YOLOv5n yet with less layers and GhostConv modules. }
    \label{fig:lightyolo}
\end{figure}

\begin{figure}[t]
    \centering
    \includegraphics[width=0.8\textwidth]{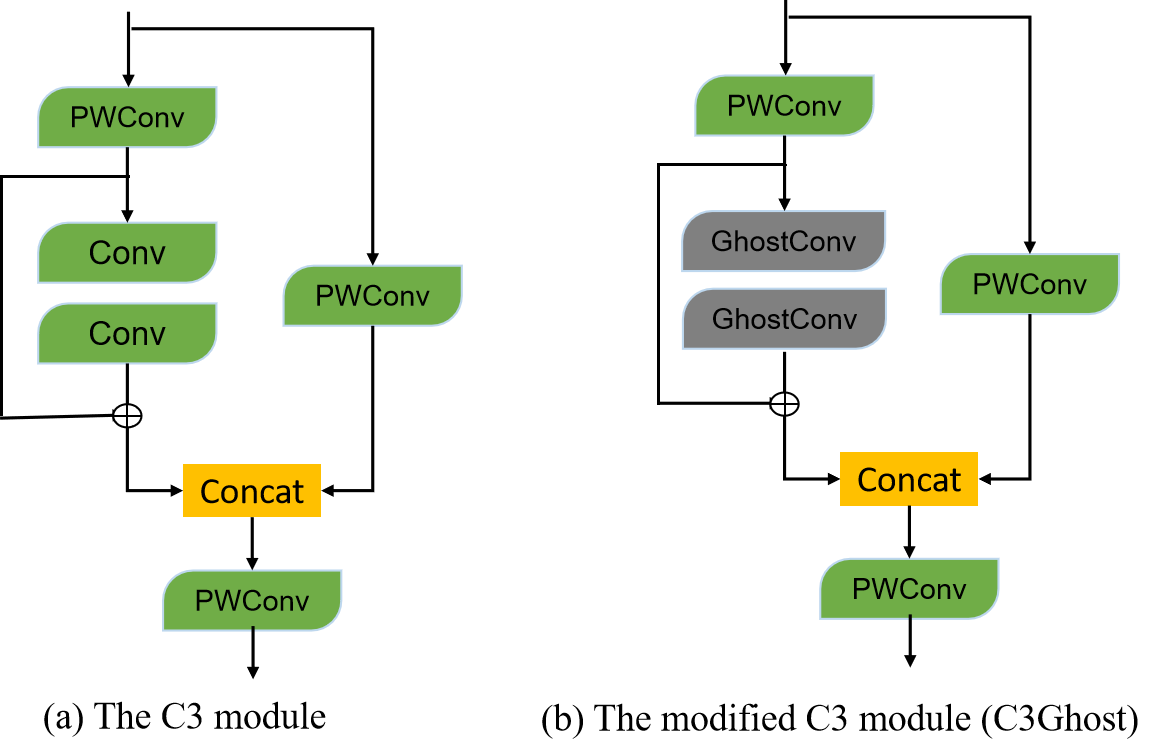}
    \caption{Our light-weight YOLOv5 vehicle smoke detector.}
    \label{fig:c3comparison}
\end{figure}
\subsection{Light-weight smoke detector}
As the first stage of our CoDeS, the smoke detector should make a trade off between speed and accuracy. It is well-known that YOLO framework is one of the most famous object detection algorithms due to its speed and accuracy. Thus, we choose YOLO as the coarse smoke detector of our CoDeS framework. Since the basic YOLOv1~\cite{2016You} framework, there have been five versions and the recent one is YOLOv5.  In YOLOv5, there exists several model structures which have different computation cost and parameters, and YOLOv5n (Nano) is the smallest one with 1.9M parameters and about 4.5 GFLOPs which can be run in real-time with CPU solutions. 

Our light-weight vehicle smoke detector is transferred from the  YOLOv5n with even less computation cost.  Figure \ref{fig:lightyolo} illustrates its structure. Specifically, our light-weight vehicle smoke detector first decrease the depth of YOLOv5n by fixing the number of C3 module as one, and then replace the C3 module with a so-called C3Ghost module inspired by MobileNet~\cite{2018MobileNetV2}  and GhostNet~\cite{2020GhostNet}. Figure \ref{fig:c3comparison} shows the comparison between C3 and C3Ghost. As it is known in MobileNet, splitting a traditional convolution operation into a depth-wise and a point-wise convolution can save computation cost and parameters largely. The GhostNet demonstrates that the redundancy in feature maps (i.e. ghost features) can be obtained by a series of cheap linear transformations, resulting in a new GhostConv module. Supposing we set $n$ output feature maps, the standard GhostConv module first uses $m (m<n)$ traditional convolutional kernels with non-linear activation function to generate $m$ feature maps, and then utilizes depth-wise convolution without non-linear operations to obtain other $s$ ghost feature maps, where $m+s=n$. We replace the traditional convolutional layers of C3 module with the plug-and-play GhostConv module, and get the C3Ghost module.
With these changes, our smoke detector only has \textit{1.2M parameters and 2.8 GFLOPs which reduces the computation cost of YOLOv5n significantly.}

\subsection{Smoke-vehicle matching}

For the pure YOLOv5 smoke detector, false positives are inevitable since it may return all the smoke-like regions. These false positives are serious especially on wet roads, shadow roads, cluttered backgrounds, etc. Worse more, these false positives are hardly to remove through frame-wise smoke detection. Considering that each vehicle smoke is emitted from a certain vehicle, we introduce a smoke-vehicle matching strategy to refine.

\begin{figure}[t]
    \centering
    \includegraphics[width=\textwidth]{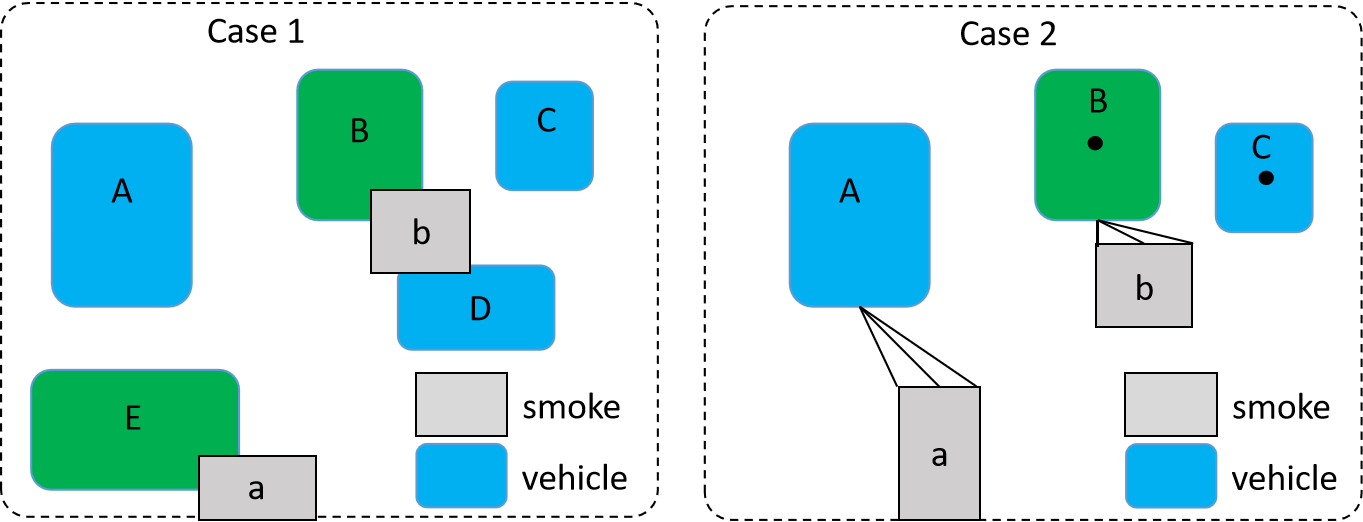}
    \caption{Two common cases in our smoke-vehicle matching strategy. The case 1 refers to that smoke regions overlap with vehicles, while the case 2 refers to that there is no overlap between smoke regions and vehicles. Green boxes denote smoky vehicles.}
    \label{fig:smokevehicle}
\end{figure}

Figure \ref{fig:smokevehicle} depicts our smoke-vehicle matching strategy. Once a smoke region is detected, we resort to the YOLOv5n model pretrained on COCO data to detect vehicles. The vehicles include car, bus, and truck in the COCO dataset. We also have tried to train a vehicle detector on our own vehicle data but found it is less robust than the pretrained one. After vehicle detection, we obtain both the vehicle regions and smoke regions. As shown in Figure \ref{fig:smokevehicle}, there would be two cases that need us to deal with separately. For both cases, considering the smoke region usually comes from the back of vehicles, we only consider the front vehicles of a certain smoke region as possible smoky vehicles, i.e., the vehicle with its center in the above of the smoke center. The first case is that the smoke region is overlapped with vehicle regions. For this case, our strategy regards the largest overlapped vehicle as the smoky vehicle, see the green regions in the left of Figure \ref{fig:smokevehicle}.  For the second case that there is no overlapped vehicles, we first compute the Euclidean distances between the bottom-middle points of vehicles and the top-left (also top-middle and top-right) points of smoke regions, and then we set a threshold $L_{dist}$ for the mean distance. A smoke-vehicle pair keeps the smallest mean distance that is less than $L_{dist}$, see the green regions in the right of Figure \ref{fig:smokevehicle}. 

\textbf{Discussion}. One could argue for the first case that it would make mistakes if there is a traffic jam. This is the truth in practice but still we have captured vehicle smoke and maybe a further check from human is needed. For the second case, the reason why we do not just use the center points between vechiles and smoke regions for distance computation is that smoky vehicles usually are those large trucks whose center points can be far from smoke regions.

\begin{figure}[t]
    \centering
    \includegraphics[width=\textwidth]{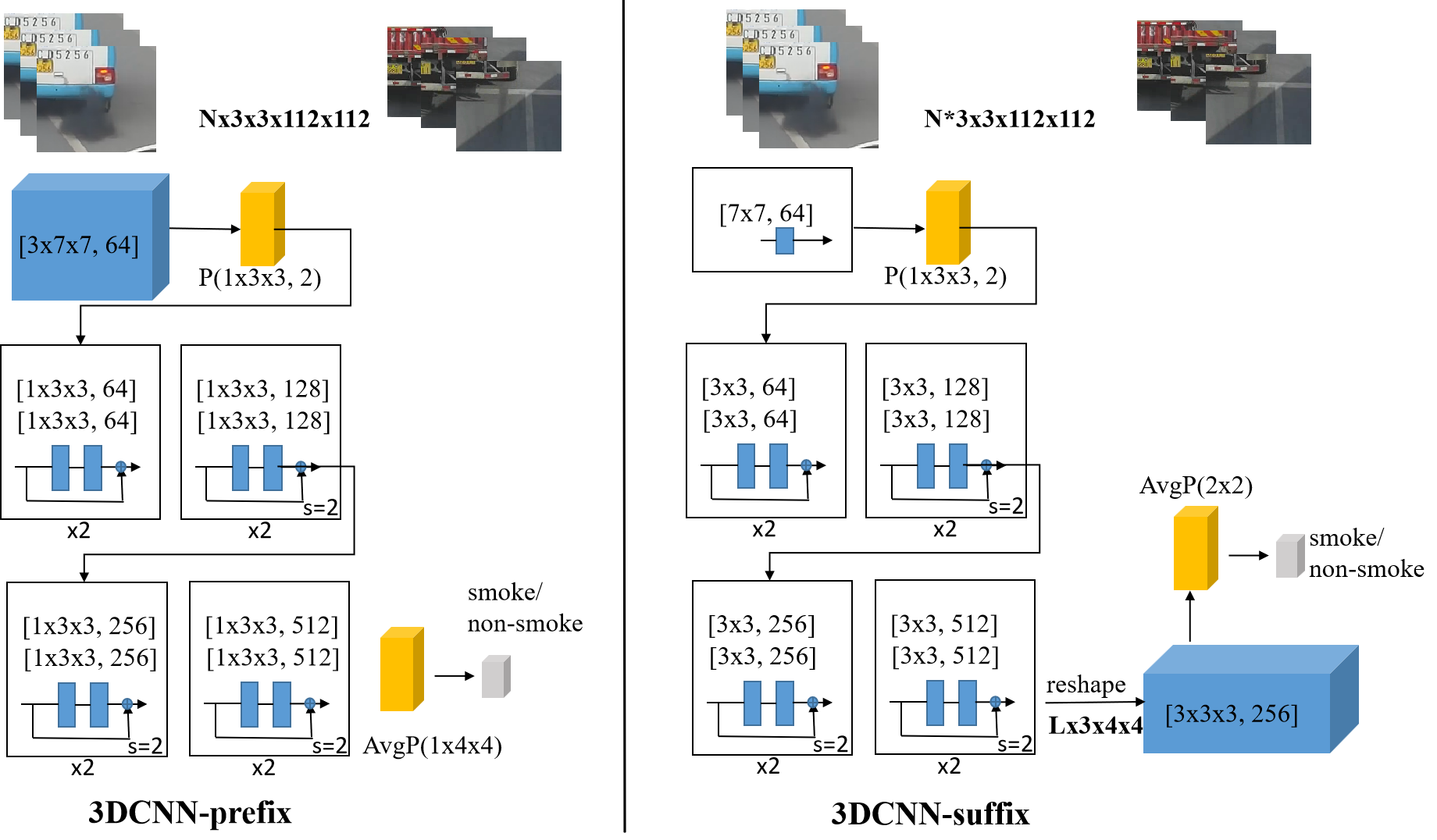}
    \caption{Our 3D convolutional networks for spatial-temporal smoke recognition.}
    \label{fig:smoke3d}
\end{figure}
\subsection{Light-weight 3D convolutional network}
After the smoke-vehicle matching stage, those detections that cannot connect to a certain vehicle are removed. However, those false positives from vehicle shadow regions can not be filtered effectively. We find that those vehicle shadow regions are even hard for human being if only the region of a frame is available.
 However, when we check the original video stream, we find it is possible to filter these false positives by non-expert people. We argue the reason could be that more information is available in spatio-time, see the top of Figure \ref{fig:smoke3d}.

Inspired by the above fact, we extend the detected regions to larger square ones and design two kinds of light-weight 3D convolutional neural networks (3DCNN) to handle those shadow regions. Specifically, if the largest side of a detected region is less than 112, we extend it to a 112$\times$112 one sharing the same center. Otherwise, we extend a detected region to a square region with the side as the largest side of the detected region. We also extend a detected frame region in time space, i.e. cropping the same region in $K$ (e.g. 3, 5, 9, etc) frames. For the first 3DCNN model, which is called 3DCNN-prefix, we take the ResNet18 as a template and adjust the first several layers to 3D convolutional layers according to the $K$, see the left of Figure \ref{fig:smoke3d} when K=3. For the second 3DCNN model in the right of Figure \ref{fig:smoke3d}, we add a 3D convolutional layer after the last 2D convolutional layer according to the $K$, and then conduct adaptive pooling. We call this architecture as 3DCNN-suffix. Intuitively, 3DCNN-prefix may be better in capturing low-level smoke motion features, while 3DCNN-suffix keeps static high-level abstract features for further motion analysis. An evaluation will be made in the next section.



\section{Experiments}
\label{experiment}
\subsection{Data splits and metrics}
Our LaSSoV image data contains 75,000 annotated smoke images, and we randomly split it into a training set and a testing set with 70,000 images and 5,000 images, respectively. We compare our YOLOv5tiny smoke detector to others on the testing set. 
For our LaSSoV-video data, we also randomly split it into a training set and a testing set with 82 videos and 81 videos. The testing set totally contains 6,727 smoky frames which is slightly more than the one of the private SEU-SmokeVeh\cite{Tao2019a}. Since existing methods did not mention the accuracy of smoke bounding boxes (i.e. IoU), we compare our method to others on the testing set of LaSSoV-video in frame level performance.

\textbf{Metrics.} For smoke detector evaluation, following popular detection metrics, we use mAP@0.5 IoU (Intersection over Union) and mAP@0.5:0.95 IoU for performance evaluation.
For smoky vehicle detection, according to existing works, we can evaluate a method in frame level, vehicle level, and segment (or video) level. Following \cite{Tao2019a,TAO202017}, we use detection rate (DR) and false alarm rate (FAR) in frame level. Specifically, the formulas of them are as follows,
\begin{equation}
    DR = \frac{TP}{N_{pos}}, ~~~ FAR=\frac{FP}{N_{neg}},
\end{equation}
where TP denotes true positve (i.e. smoky frame) and FP is false positive, $N_{pos}$ and $N_{neg}$ are the numbers of positives and negatives. \textit{It is worth noting that DR is called as the probability of correct classification on smoky frames, and also that DR is identical to the Recall conception}. In addition, we also use Precision and F1 score for frame-level performance evaluation which is also important for algorithm comparison. Their formulas are as follows,
\begin{equation}
    Precision = \frac{TP}{TP+FP}, ~~~ F1=\frac{2\times Precision \times Recall}{ Precision + Recall}.
\end{equation}

\subsection{Implementation details}
We use the PyTorch toolbox\footnote{https://pytorch.org/} to implement our method on a Linux server with two GeForce RTX 3090 GPUs. Specifically, YOLO based smoke detectors are implemented with the open source code\footnote{https://github.com/ultralytics/yolov5}, and are trained from scratch. We set initial learning rate to 0.01, batch size to 64, and training epochs to 30. All other parameters are set by default. The threshold $L_{dist}$ of the smoke-vehicle matching module is set by default to 50 which means the distance between a smoke region and a vehicle should be less than 50 pixels.

For the training data of light-weight 3DCNN models, we first apply our YOLOv5tiny smoke detector (with score threshold 0.2) and smoke-vehicle matching module on LaSSoV-video training set to collect sequence samples. The temporal extent $K$ is set by default to 3, and 10,033 sequences are finally obtained to train 3DCNN models. For data augmentation, we resize the crop regions in sequences to 128$\times$128 and then randomly crop 112$\times$112 regions for training. We set initial learning rate to 0.01, and decrease it by a factor 0.1 after every 4 epochs, and stop training at the 10-th epoch.

\begin{table}[t]
\caption{Evaluation of different deep learning detectors on LaSSoV. All the detectors are trained on the training set of LaSSoV and the results are reported on the test set of LaSSoV with image size 640. The speeds are reported on three randomly-selected smoke images with average inference time.}
\resizebox{\columnwidth}{!}{
\begin{tabular}{p{2.4cm}ccccp{1.5cm}p{1.5cm}}
\toprule
Detector         & mAP@0.5 & mAP@0.5:0.95 & Params(M) & FLOPs &Speed-CPU(ms) & Speed-GPU(ms) \\ \midrule
MobileNetv2-YOLOv5s &   64.52   &    22.96     &  4.6   &    36.7   &   470.6 &   13.1      \\
YOLOv5s          &    87.3  &    39.8     &   7.2   &    16.5   &   134.5 &   18.5      \\
YOLOv5n          &    80.9     &   33.1         &   1.9      &   4.5    &   61.5     &   13.2           \\
Our~YOLOv5tiny &          79.8        &   32.1     &   1.2     &   2.8   &   48.9   &   12.8    \\ 
\bottomrule
\end{tabular}}
\label{tab:detector}
\end{table}

\subsection{Detector evaluation on LaSSoV}
We evaluate several variations of YOLOv5, including YOLOv5s, YOLOv5n, and MobileNetv2~\cite{2018MobileNetV2} backbone based YOLOv5s. Table \ref{tab:detector} compares their mean Average Precision and computation cost. For the computation cost, we run 3 times on 3 smoke samples and report the average cost. From Table \ref{tab:detector}, several observations can be concluded as follows. First, replacing the default YOLOv5s backbone by MobileNetv2 gets the worst results and is not efficient. It is worth noting that the recent YOLOv5 architecture has already introduced efficient modules and is better optimized than MobileNetv2. 
Second, YOLOv5s obtains the best accuracy while needs more than twice inference cost compared to YOLOv5n and our YOLOv5tiny. Finally, compared to YOLOv5n, our model obtains similar performance but has less FLOPs, and is more efficient in CPU due to the Ghost module.

\subsection{Comparison to the state of the art on LaSSoV-video}
As above mentioned, our LaSSoV-video contains diverse scenes which could be challenging for existing methods. We evaluate the pure YOLOv5s, our YOLOv5tiny, and our CoDeS on the test set (4 scenes) of LaSSoV-video and report the average performance.  The detection threshold is set to 0.2 for our CoDeS. For the pure YOLOv5s and YOLOv5tiny, we report the results with detection thresholds both 0.2 and 0.5. We re-implement the R-VLBC~\cite{Tao2019a} on our LaSSoV-video and also compare to other methods on the private SEU-SmokeVeh (results are copied from the reference papers). Unlike existing works which only use one or two metrics, we apply four popular frame-wise metrics for evaluation.

\begin{table}[t]
\centering
\caption{Comparison between our method and state-of-the-art methods with varied metrics. The first four methods are evaluated on our LaSSoV-video, and the others are reported in existing papers on the private SEU-SmokeVeh (Dataset 5)~\cite{Tao2019a} which is similar with ours in data scale. The values in parentheses denotes the detection thresholds of YOLOv5. The time is reported on CPU for one frame.}
\begin{tabular}{lclllll}
Method   & DR & FAR($\downarrow$) & Precision & F1 & Time/ms \\\toprule
YOLOv5s(0.2)  &  0.9368  &  0.1349   & 0.4064    &  0.5580  & 134.5\\
YOLOv5s(0.5)   &  0.8471   &  0.0453   & 0.6379    &  0.7206  &  134.5 \\
YOLOv5tiny(0.2) &  0.9087  &  0.1436   &    0.3967       &  0.5370  & 48.9\\ 
YOLOv5tiny(0.5)  &  0.7168   &  0.0287    &   0.7000       &  0.6949  & 48.9\\ 
Our CoDeS(3DCNN-prefix) &  0.7381&	0.0297&	0.7045&	0.7200 & 68.2+61.5\\ 
Our CoDeS(3DCNN-suffix) &  0.7713&	0.0192&	0.7936&\textbf{0.7801}& 72.2+61.5\\ 
R-VLBC~\cite{Tao2019a}  & 0.8453   &  0.1893   &     0.3171	 & 0.4623   &160.56\\ \midrule
R-VLBC~\cite{Tao2019a}   &0.9253   &  0.0893   &     --      & --   & \\ 
LBP~\cite{TAO202017} & 0.8674   &  0.3103   &   --        &  --  &--\\  
NR-RLBP~\cite{TAO202017}& 0.9125    &   0.1610  &    --       &  -- & 140.34\\ 
RF-TOP~\cite{Tao2018a}  & 0.8757   &  0.1236   &     --      & --   &--\\ 
\bottomrule
\end{tabular}
\label{tab:eval_codes}
\end{table}

\begin{figure}[t]
    \centering
    \includegraphics[width=0.9\textwidth]{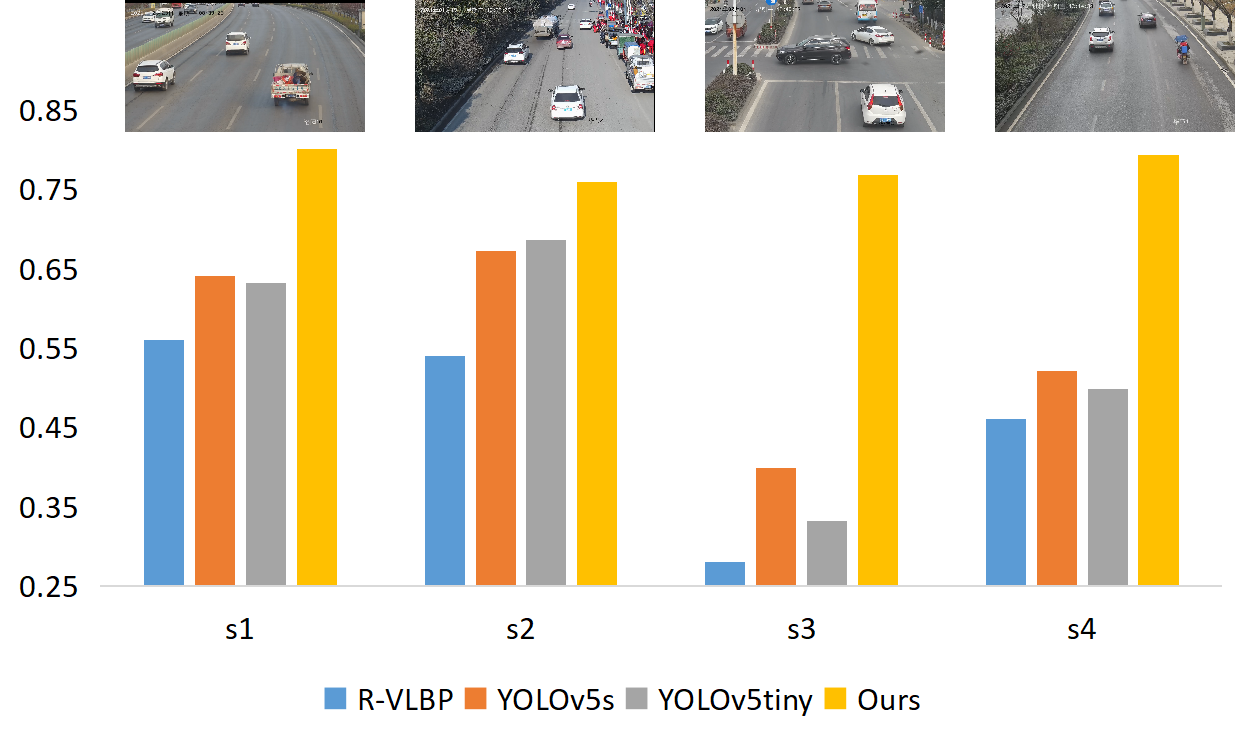}
    \caption{The F1 scores on four scenes. To clarify, we illustrate a sample frame for each scene on the top of figure. }
    \label{fig:f1eval}
\end{figure}

\begin{figure}[t]
    \centering
    \includegraphics[width=\textwidth]{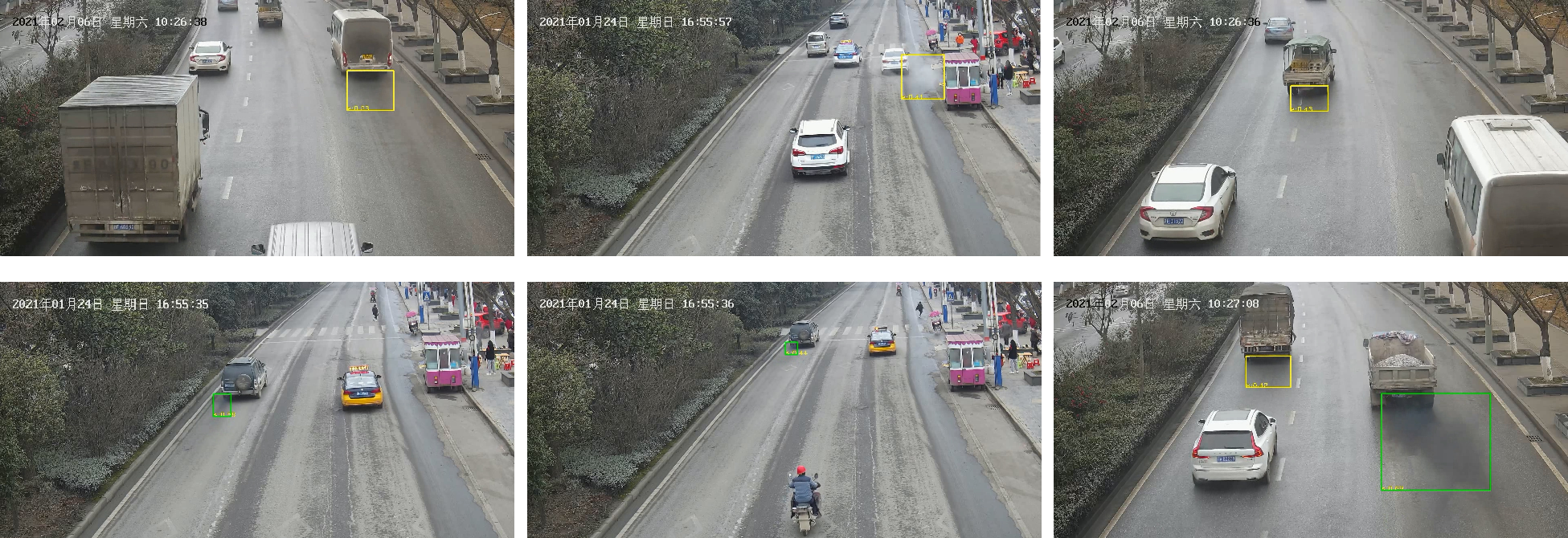}
    \caption{Detection samples in LaSSoV-video test set. Yellow bounding boxes denote those false positives of YOLOv5tiny, and green bounding boxes refer to those results of our CoDeS. In other words, the yellow boxes are filtered by the V-S matching module and 3D CNN module of CoDeS.}
    \label{fig:samples}
\end{figure}

From Table \ref{tab:eval_codes}, several observations can be concluded as follows. First, from the first and the third rows, we find that YOLOv5s obtains the highest detection ratio (or recall ratio) while the performance gap between YOLOv5s and our YOLOv5tiny is not as large as the one in the LaSSoV image data. This suggests the light-weight YOLOv5tiny is enough to learn the main features of obvious smoke regions while YOLOv5s can struggle to hard cases. Second, increasing the detection threshold to 0.5 for YOLOv5s and YOLOv5tiny largely improves precision and degrades detection rates, while still boosts F1 scores. Third, 3DCNN-suffix is consistently superior to 3DCNN-prefix for our CoDeS framework, which may be explained by that motion information in high-level layers is better for classification. \textit{We fix 3DCNN-suffix as the 3DCNN module of our CoDeS in the remaining content unless otherwise specified.} Forth, our CoDeS method significantly outperforms the others in Precision and F1, while still keeps reasonable detection rates. Specifically, based on the results of YOLOv5tiny(0.2), our method obtains around 25\% gain in F1 score. Last but not the least, the R-VLBC~\cite{Tao2019a} method that leverages hand-crafted local pattern features gets the worst performance in all metrics, which indicates the limitations of these features.

To further investigate their performance in different scenes, Figure \ref{fig:f1eval} illustrates the comparison of F1 score in each scene. The traditional R-VLBC method is consistently worse than deep learning methods especially in scene \#3 (cross roads). There exists a lots of varied vehicle shadows in scene \#3, which largely confuses those static smoke detectors. YOLOv5s(0.2) and YOLOv5tiny(0.2) perform good and similarly in simple scenes (i.e., scene \#1 and \#2). Scene \#4 is partially wet on the road whose wet regions can be easily mistaken as smoke regions by YOLOv5s and YOLOv5tiny, leading to low F1 scores. Our CoDeS is significantly superior to other methods in all scenes thanks to the 3DCNN classifier. Specifically, our CoDeS improves the F1 of YOLOv5s by around 43\% in scene \#3, indicating that a 3DCNN is robust to different roads. 

Figure \ref{fig:samples} further illustrates some detection samples in our LaSSoV-video test set. The yellow bounding boxes denote those false positives of YOLOv5tiny with scores larger than 0.4. These false positives are even hard for human beings in static images, which are either classified to non-smoke category in our 3DCNN-suffix model or filtered out by our S-V matching module (the top-middle one).

\textbf{Analysis on computational cost}. In Table \ref{tab:eval_codes}, we also report the computational cost on CPU (Intel Xeon E5-2678 v3) for processing one frame. Specifically, the cost of our CoDeS depends on the YOLOv5tiny, the 3DCNN, and the extra YOLOv5n for vehicle detection (61.5ms). It is worth noting that all other methods did not provide vehicle regions yet it is necessary for practical applications. In general, our CoDeS is efficient on CPU platform which obtains 78.01\% F1 score with speed around 8 FPS (frames per second). 

\begin{table}[t]
\centering
\caption{Evaluation of the individual modules of our CoDeS on LaSSoV-video test set. }
\resizebox{\columnwidth}{!}{
\begin{tabular}{ccc|cccc}
\toprule 
 YOLOv5tiny & S-V matching & 3DCNN-suffix & DR & FAR($\downarrow$) & Precision & F1 \\ \midrule
 $\surd$ &    &       &       0.9087  &  0.1436   &    0.3967       &  0.5370      \\
  $\surd$ &  $\surd$  &    &     0.8815  &  0.1211   &  0.4255   &   0.5616 \\
  $\surd$ &  $\surd$  &  $\surd$  &     0.7713&	0.0192&	0.7936& 0.7801
  \\\bottomrule
\end{tabular}}
\label{tab:modules}
\end{table}

\begin{figure}[t]
    \centering
    \includegraphics[width=0.9\textwidth]{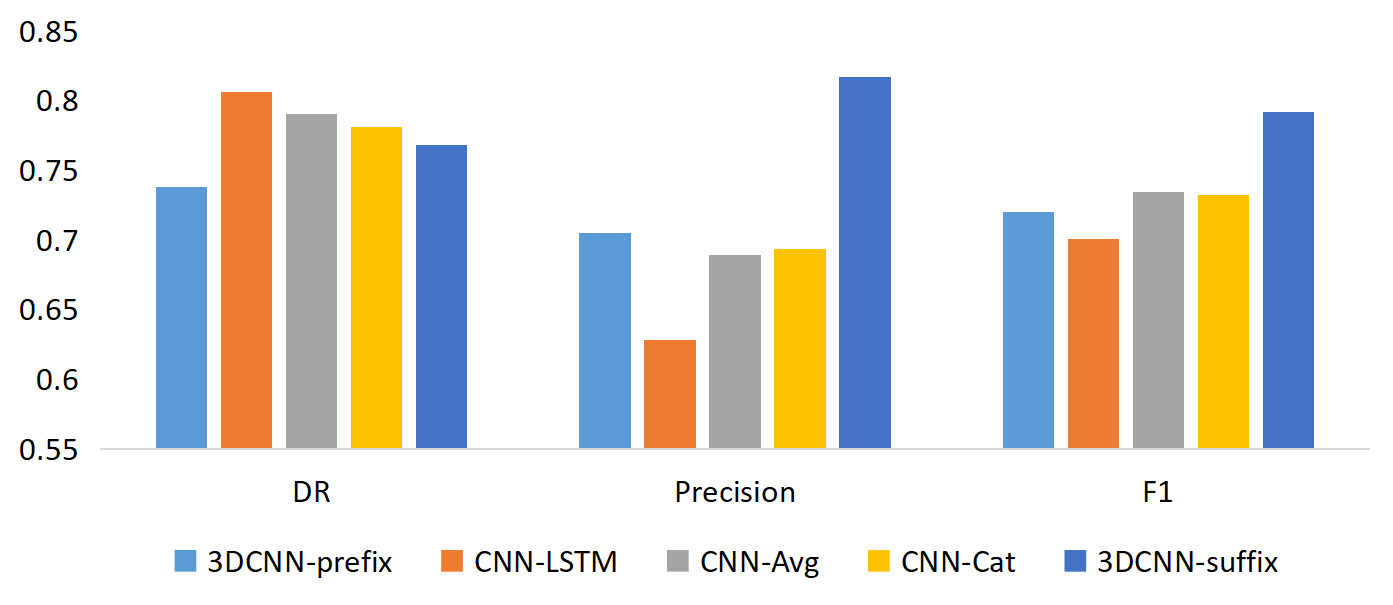}
    \caption{The comparison of different spatial-temporal models. }
    \label{fig:stmodel}
\end{figure}



\begin{figure}[t]
    \centering
    \includegraphics[width=0.32\textwidth]{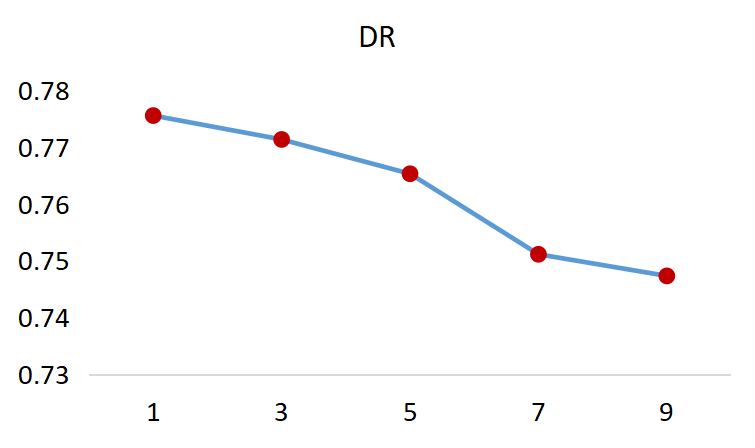}
    \includegraphics[width=0.32\textwidth]{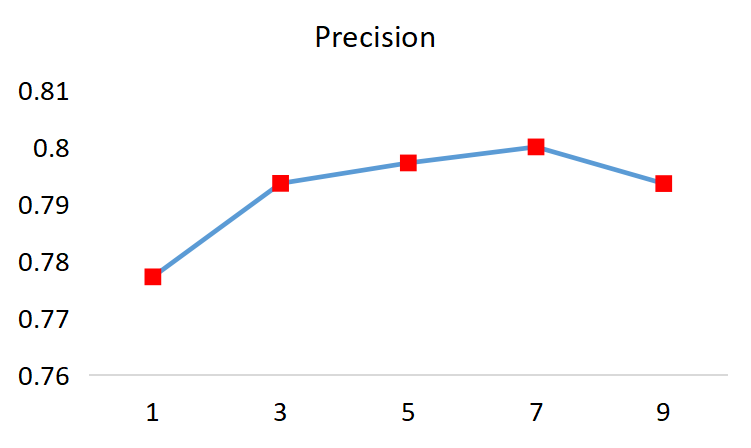}
    \includegraphics[width=0.32\textwidth]{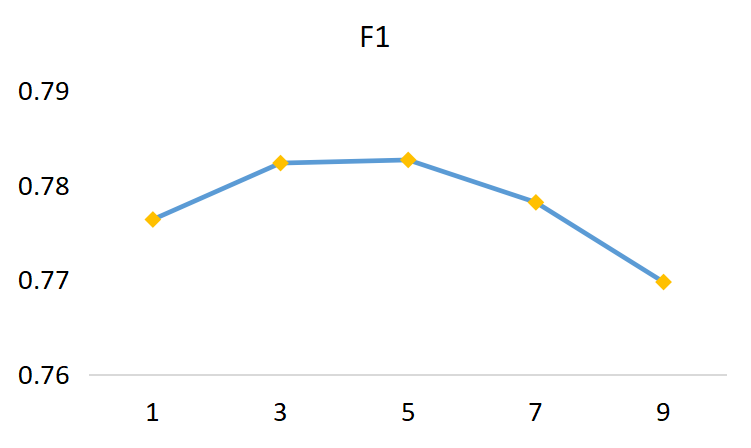}
    \caption{Evaluation of temporal length for our 3DCNN-suffix.}
    \label{fig:evalLen}
\end{figure}
\subsection{Ablation study}
\textbf{The smoke-vehicle matching and 3DCNN modules}. Since our CoDeS framework including two auxiliary modules compared to pure YOLO detectors, we conduct an evaluation for these two modules in Table \ref{tab:modules}. From Table \ref{tab:modules}, the S-V matching module slightly improves YOLOv5tiny in almost all metrics except for the detection rate (a slight degradation), and the 3DCNN module brings larger performance gains than the S-V matching module. Specifically, for the F1 scores, the S-V matching module contributes 2.5\% improvement while the 3DCNN module contributes 22\%.

\textbf{Spatial-temporal models}.
We evaluate several other common spatial-temporal video classification models including long-short term memory with CNN features (CNN-LSTM)~\cite{2015Beyond}, temporal segment network~\cite{2016Temporal} (i.e. temporal average with CNN features, CNN-Avg), and temporal concatenation with CNN features (CNN-Cat). We replace our 3DCNN-suffix as one of them for the evaluation. All the backbones of them are the same as our 3DCNN model (i.e., ResNet-18), the CNN features are the output of the last Pooling layer in ResNet-18. The results are shown in Figure \ref{fig:stmodel}. Several observations can be concluded as follows. First, our 3DCNN-suffix obtains the best F1 score and Precision which suggests the abstract motion information may be captured by 3D convolutions. Second, CNN-Avg and CNN-Cat perform similarly in all metrics since they actually use the same static CNN features of three frames. Last but not the least, a LSTM model upon CNN features keeps high recall but is not effective in distinguishing smoke regions.

\begin{figure}[t]
    \centering
    \includegraphics[width=\textwidth]{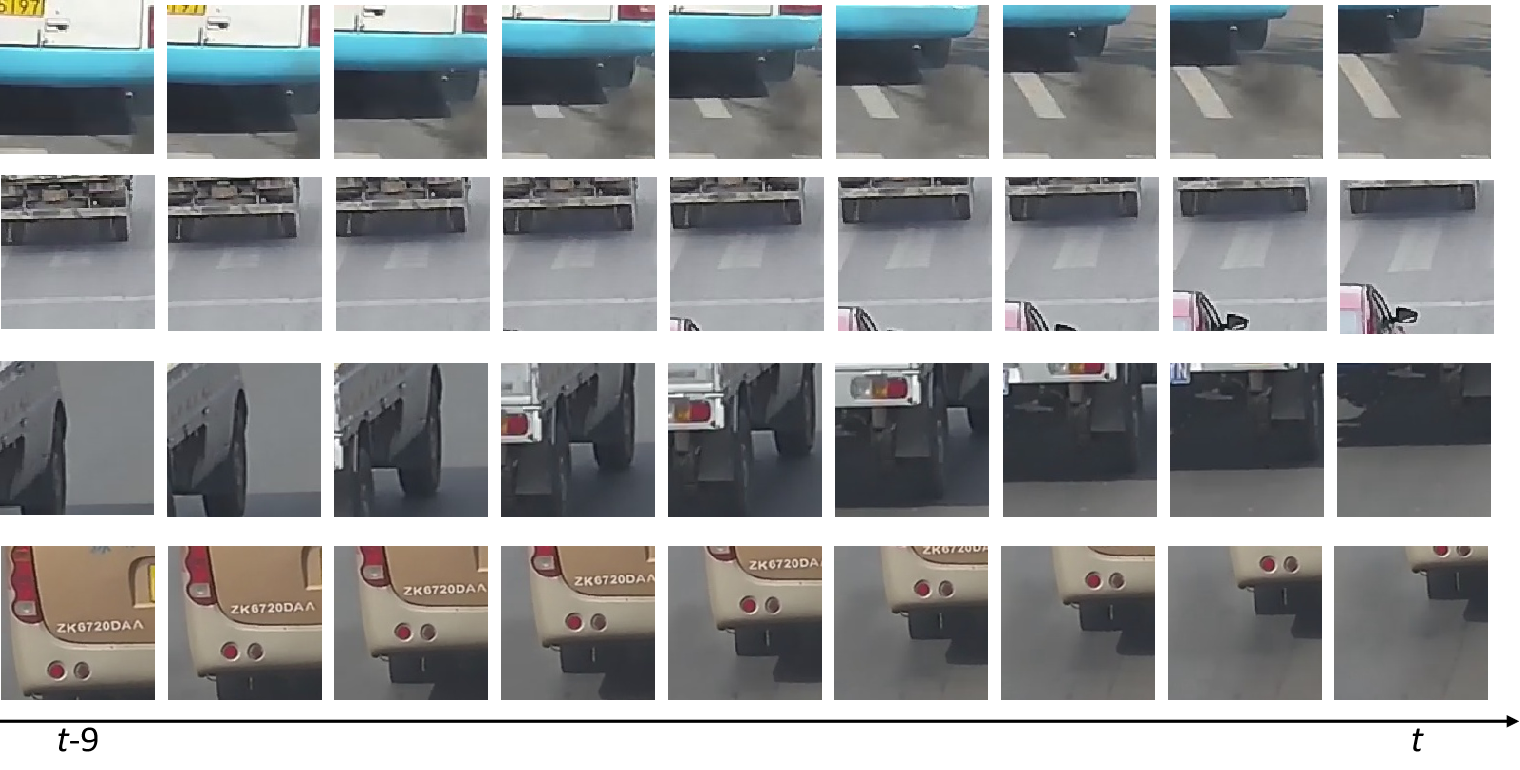}
    \caption{Temporal extent examples for detected smoke regions in time \textit{t}.}
    \label{fig:examLen}
\end{figure}

\textbf{The input length of 3DCNN}. 
As mentioned above, the default temporal input length is 3 for our 3DCNN modules. In fact, this short temporal extent is unusual in other similar tasks like video action recognition\footnote{https://pytorchvideo.org/}~\cite{2015Beyond,2015Learning,2016Temporal}. Thus, we make an evaluation on the temporal length as shown in Figure \ref{fig:evalLen}. We do not show the false alarm rates since all of these models get less than 0.02 in FAR. We take ResNet-18 to finetune an image classification model for length 1. As can be seen in Figure \ref{fig:evalLen}, the F1 score and precision are improved when increasing the length but saturated at length 3. The detection rate is slightly degraded with increasing input length. We show some temporal extent samples in Figure \ref{fig:examLen} to investigate why more frames is not helpful for performance. We find that most of the extent frame regions are not related to smoke, which is mainly caused by the high-speed vehicles and low frame rate cameras.

\textbf{The input size of 3D CNN}.
As known in image classification, large input size is helpful for performance. To verify it, we rescale the input to 256$\times$256, and randomly crop 224$\times$224 regions to train our 3DCNN-suffix model with temporal length 3. Replacing the default 3DCNN-suffix by this one, we achieve 0.7552, 0.0176, 0.8037, and 0.7763, in DR, FAR, Precision, and F1. Compared to the results in Table \ref{tab:eval_codes}, we find that only Precision is slightly improved while there is a obvious degradation in DR or F1. This may be explained by that most of the cropping `smoke' regions are smaller than 224 since smoky vehicles are far away from cameras, and zoom-in regions can disturb the smoke characteristics.



\section{Conclusion and future work}
\label{conclusion}
In this paper, we explored the issue of smoky vehicle detection. We first built a large-scale smoky vehicle image dataset, LaSSoV, which is excepted to use for training deep learning models. As there is no public smoky vehicle test videos, we also provided a video dataset called LaSSoV-video for algorithm comparison. Moreover, we presented a new Coarse-to-fine Deep Smoky vehicle detection (CoDeS) framework for smoky vehicle detection, which is demonstrated to be efficient on CPU platform. In addition, we comprehensively evaluated our method and others in four metrics which can be more fair than only in one or two. As found in our 3DCNN module, there are many irrelevant regions when increasing the temporal length, we think a tracking process after detecting can be better which is our future work.

\bibliography{mybibfile}

\begin{thebibliography}{10}
\expandafter\ifx\csname url\endcsname\relax
  \def\url#1{\texttt{#1}}\fi
\expandafter\ifx\csname urlprefix\endcsname\relax\def\urlprefix{URL }\fi
\expandafter\ifx\csname href\endcsname\relax
  \def\href#1#2{#2} \def\path#1{#1}\fi

\bibitem{science1995}
S.~P. Beaton, G.~A. Bishop, Y.~Zhang, D.~H. Stedman, L.~L. Ashbaugh, D.~R.
  Lawson, On-road vehicle emissions: Regulations, costs, and benefits, Science
  268~(5213) (1995) 991--993.

\bibitem{2009Real}
K.~Ropkins, J.~Beebe, H.~Li, B.~Daham, J.~Tate, M.~Bell, G.~Andrews, Real-world
  vehicle exhaust emissions monitoring: Review and critical discussion,
  Critical Reviews in Environmental Science and Technology 39~(2) (2009) 79--152.

\bibitem{TAO202017}
H.~Tao, P.~Zheng, C.~Xie, X.~Lu, A three-stage framework for smoky vehicle
  detection in traffic surveillance videos, Information Sciences 522 (2020)
  17--34.

\bibitem{Tao2018}
H.~Tao, X.~Lu, {Smoky vehicle detection based on multi-scale block Tamura
  features}, Signal, Image and Video Processing 12~(6) (2018) 1061--1068.

\bibitem{Cao2018}
Y.~Cao, C.~Lu, X.~Lu, X.~Xia, {A Spatial Pyramid Pooling Convolutional Neural
  Network for Smoky Vehicle Detection}, in: 2018 37th Chinese Control
  Conference (CCC), IEEE, 2018, pp. 9170--9175.

\bibitem{Tao2019b}
H.~Tao, X.~Lu, {Automatic smoky vehicle detection from traffic surveillance
  video based on vehicle rear detection and multi-feature fusion}, IET
  Intelligent Transport Systems 13~(2) (2019) 252--259.

\bibitem{Tao2019a}
H.~Tao, X.~Lu, {Smoke vehicle detection based on robust codebook model and
  robust volume local binary count patterns}, Image and Vision Computing 86
  (2019) 17--27.

\bibitem{Cao2019}
Y.~Cao, X.~Lu, {Learning spatial-temporal representation for smoke vehicle
  detection}, Multimedia Tools and Applications 78~(19) (2019) 27871--27889.

\bibitem{Tao2019}
H.~Tao, X.~Lu, {Smoke Vehicle Detection Based on Spatiotemporal Bag-Of-Features
  and Professional Convolutional Neural Network}, IEEE Transactions on Circuits
  and Systems for Video Technology 30~(10) (2019) 3301--3316.

\bibitem{Tao2020}
H.~Tao, X.~Lu, {Smoke vehicle detection based on multi-feature fusion and
  hidden Markov model}, Journal of Real-Time Image Processing 17~(3) (2020)
  745--758.

\bibitem{Wang2021}
C.~Wang, H.~Wang, F.~Yu, W.~Xia, {A High-Precision Fast Smoky Vehicle Detection
  Method Based on Improved Yolov5 Network}, in: 2021 IEEE International
  Conference on Artificial Intelligence and Industrial Design (AIID), IEEE,
  2021, pp. 255--259.

\bibitem{toreyin2005wavelet}
B.~U. T{\"o}reyin, Y.~Dedeo{\u{g}}lu, A.~E. Cetin, Wavelet based real-time
  smoke detection in video, in: 2005 13th European signal processing
  conference, IEEE, 2005, pp. 1--4.

\bibitem{xiong2007video}
Z.~Xiong, R.~Caballero, H.~Wang, A.~M. Finn, M.~A. Lelic, P.-Y. Peng,
  Video-based smoke detection: possibilities, techniques, and challenges, in:
  IFPA, fire suppression and detection research and applications—a technical
  working conference (SUPDET), Orlando, FL, 2007.

\bibitem{yuan2008fast}
F.~Yuan, A fast accumulative motion orientation model based on integral image
  for video smoke detection, Pattern Recognition Letters 29~(7) (2008)
  925--932.

\bibitem{ZHOU201650}
Z.~Zhou, Y.~Shi, Z.~Gao, S.~Li, Wildfire smoke detection based on local
  extremal region segmentation and surveillance, Fire Safety Journal 85 (2016)
  50--58.

\bibitem{2017Smoke}
G.~Lin, Y.~Zhang, Q.~Zhang, Y.~Jia, J.~Wang, Smoke detection in video sequences
  based on dynamic texture using volume local binary patterns, Ksii
  Transactions on Internet and Information Systems 11~(11) (2017) 5522--5536.

\bibitem{yuan2018learning}
F.~Yuan, X.~Xia, J.~Shi, L.~Zhang, J.~Huang, Learning multi-scale and
  multi-order features from 3d local differences for visual smoke recognition,
  Information Sciences 468 (2018) 193--212.

\bibitem{yuan2019encoding}
F.~Yuan, J.~Shi, X.~Xia, L.~Zhang, S.~Li, Encoding pairwise hamming distances
  of local binary patterns for visual smoke recognition, Computer Vision and
  Image Understanding 178 (2019) 43--53.

\bibitem{2019Convolutional}
F.~Yuan, L.~Zhang, B.~Wan, X.~Xia, J.~Shi, Convolutional neural networks based
  on multi-scale additive merging layers for visual smoke recognition, Machine
  Vision and Applications 30~(2) (2019) 345--358.

\bibitem{yuan2019fusing}
F.~Yuan, G.~Li, X.~Xia, B.~Lei, J.~Shi, Fusing texture, edge and line features
  for smoke recognition, IET Image Processing 13~(14) (2019) 2805--2812.

\bibitem{2020Encoding}
F.~Yuan, G.~Li, X.~Xia, J.~Shi, L.~Zhang, Encoding features from multi-layer
  gabor filtering for visual smoke recognition, Pattern Analysis and
  Applications 23~(3) (2020) 1117--1131.

\bibitem{YUAN20124326}
F.~Yuan, A double mapping framework for extraction of shape-invariant features
  based on multi-scale partitions with adaboost for video smoke detection,
  Pattern Recognition 45~(12) (2012) 4326--4336.

\bibitem{1998Statistical}
V.~N. Vapnik, Statistical learning theory, Encyclopedia of the Sciences of
  Learning 41~(4) (1998) 3185--3185.

\bibitem{2012ImageNet}
A.~Krizhevsky, I.~Sutskever, G.~Hinton, Imagenet classification with deep
  convolutional neural networks, in: NIPS, 2012.

\bibitem{2017Comparative}
A.~Filonenko, L.~Kurnianggoro, K.~H. Jo, Comparative study of modern
  convolutional neural networks for smoke detection on image data, in:
  International Conference on Human System Interactions, 2017.

\bibitem{20183D}
X.~Li, Z.~Chen, Q.~Wu, C.~Liu, 3d parallel fully convolutional networks for
  real-time video wildfire smoke detection, IEEE Transactions on Circuits and
  Systems for Video Technology (2018) 1--1.

\bibitem{2019Smoke}
G.~Lin, Y.~Zhang, G.~Xu, Q.~Zhang, Smoke detection on video sequences using 3d
  convolutional neural networks, Fire Technology.

\bibitem{2019Deep}
K.~Gu, Z.~Xia, J.~Qiao, W.~Lin, Deep dual-channel neural network for
  image-based smoke detection, IEEE Transactions on Multimedia PP~(99) (2019)
  1--1.

\bibitem{2019Recurrent}
M.~Yin, C.~Lang, Z.~Li, S.~Feng, T.~Wang, Recurrent convolutional network for
  video-based smoke detection, Multimedia Tools and Applications 78~(1) (2019)
  237--256.

\bibitem{Yaocong2018Real}
Yaocong, Hu, Xiaobo, Lu, Real-time video fire smoke detection by utilizing
  spatial-temporal convnet features, Multimedia Tools and Applications.

\bibitem{2018Fire}
Y.~Luo, L.~Zhao, P.~Liu, D.~Huang, Fire smoke detection algorithm based on
  motion characteristic and convolutional neural networks, Multimedia Tools and
  Applications.

\bibitem{2016Deep}
K.~He, X.~Zhang, S.~Ren, J.~Sun, Deep residual learning for image recognition,
  in: 2016 IEEE Conference on Computer Vision and Pattern Recognition (CVPR),
  2016.

\bibitem{2014Very}
K.~Simonyan, A.~Zisserman, Very deep convolutional networks for large-scale
  image recognition, Computer Science.

\bibitem{2016You}
J.~Redmon, S.~Divvala, R.~Girshick, A.~Farhadi, You only look once: Unified,
  real-time object detection, in: Computer Vision and Pattern Recognition, 2016.

\bibitem{2018MobileNetV2}
M.~Sandler, A.~Howard, M.~Zhu, A.~Zhmoginov, L.~C. Chen, Mobilenetv2: Inverted
  residuals and linear bottlenecks, in: 2018 IEEE/CVF Conference on Computer
  Vision and Pattern Recognition (CVPR), 2018.

\bibitem{2020GhostNet}
K.~Han, Y.~Wang, Q.~Tian, J.~Guo, C.~Xu, Ghostnet: More features from cheap
  operations, in: 2020 IEEE/CVF Conference on Computer Vision and Pattern
  Recognition (CVPR), 2020.

\bibitem{Guillemant2001Real}
Guillemant, Philippe, Real-time identification of smoke images by clustering
  motions on a fractal curve with a temporal embedding method, Optical
  Engineering 40~(4) (2001) 554–563.

\bibitem{2002Smoke}
F.~Gomez-Rodriguez, S.~Pascual-Pena, B.~Arrue, A.~Ollero, Smoke detection using
  image processing, in: International conference on forest fire
  research;Wildland fire safety summit;ICFFR, 2002.

\bibitem{2016Convolutional}
S.~Frizzi, R.~Kaabi, M.~Bouchouicha, J.~M. Ginoux, E.~Moreau, F.~Fnaiech,
  Convolutional neural network for video fire and smoke detection, in:
  Conference of the IEEE Industrial Electronics Society, 2016, pp. 877--882.

\bibitem{1998Gradient}
Y.~Lecun, L.~Bottou, Gradient-based learning applied to document recognition,
  Proceedings of the IEEE 86~(11) (1998) 2278--2324.

\bibitem{2016Smoke}
C.~Tao, Z.~Jian, W.~Pan, Smoke detection based on deep convolutional neural
  networks, in: 2016 International Conference on Industrial Informatics -
  Computing Technology, Intelligent Technology, Industrial Information
  Integration (ICIICII), 2016.

\bibitem{szegedy2016rethinking}
C.~Szegedy, V.~Vanhoucke, S.~Ioffe, J.~Shlens, Z.~Wojna, Rethinking the
  inception architecture for computer vision, in: Proceedings of the IEEE
  conference on computer vision and pattern recognition, 2016, pp. 2818--2826.

\bibitem{Ko2012Wildfire}
Ko, ByoungChul, Wildfire smoke detection using temporospatial features and
  random forest classifiers, Optical Engineering 51~(1) (2012) 7208.

\bibitem{csvt2015_fire}
P.~Foggia, A.~Saggese, M.~Vento, Real-time fire detection for video
  surveillance applications using a combination of experts based on color,
  shape and motion, IEEE Transactions on Circuits and Systems for Video
  Technology.

\bibitem{YUAN2016225}
F.~Yuan, J.~Shi, X.~Xia, Y.~Fang, Z.~Fang, T.~Mei, High-order local ternary
  patterns with locality preserving projection for smoke detection and image
  classification, Information Sciences 372 (2016) 225--240.

\bibitem{Tao2018a}
H.~Tao, X.~Lu, {Smoky vehicle detection based on range filtering on three
  orthogonal planes and motion orientation histogram}, IEEE Access 6 (2018)
  57180--57190.

\bibitem{2015Beyond}
Y.~H. Ng, M.~Hausknecht, S.~Vijayanarasimhan, O.~Vinyals, G.~Toderici, Beyond
  short snippets: Deep networks for video classification, in: 2015 IEEE
  Conference on Computer Vision and Pattern Recognition (CVPR), 2015.

\bibitem{2016Temporal}
L.~Wang, Y.~Xiong, Z.~Wang, Y.~Qiao, D.~Lin, X.~Tang, L.~V. Gool, Temporal
  segment networks: Towards good practices for deep action recognition, 2016.

\bibitem{2015Learning}
D.~Tran, L.~Bourdev, R.~Fergus, L.~Torresani, M.~Paluri, Learning
  spatiotemporal features with 3d convolutional networks, in: IEEE
  International Conference on Computer Vision, 2015.

\end{thebibliography}

\end{document}